%% file: sn-article.tex
\documentclass[sn-mathphys]{sn-jnl}

\usepackage{xurl}
\usepackage{array}

\usepackage{amsmath, amssymb, amsfonts, amsthm}
\usepackage{mathtools,commath}
\usepackage{graphicx, multirow}
\usepackage{hyperref}
\usepackage{bigstrut}
\newcolumntype{H}{>{\setbox0=\hbox\bgroup}c<{\egroup}@{}}

\usepackage{amssymb}
\usepackage{pifont}
\newcommand{\cmark}{\ding{51}}%
\newcommand{\xmark}{\ding{55}}%

\jyear{2021}%

\theoremstyle{thmstyleone}%
\theoremstyle{thmstyletwo}%

\theoremstyle{thmstylethree}%

\newcommand{\TODO}[1]{{\color{red}TODO: #1}}

\newcommand{\eat}[1]{}

\begin{document}

\title[Data Acquisition: A New Frontier in Data-centric AI]{Data Acquisition: A New Frontier in Data-centric AI}



\author[1]{\fnm{Lingjiao} \sur{Chen}}\email{lingjiao@stanford.edu}
\author[2]{\fnm{Bilge} \sur{Acun}}\email{acun@meta.com}
\author[2]{\fnm{Newsha} \sur{Ardalani}}\email{new@meta.com}
\author[3]{\fnm{Yifan} \sur{Sun}}\email{ys3600@columbia.edu}
\author[4]{\fnm{Feiyang} \sur{Kang}}\email{fyk@vt.edu}
\author[3]{\fnm{Hanrui} \sur{Lyu}}\email{hl3616@columbia.edu}
\author[3]{\fnm{Yongchan} \sur{Kwon}}\email{yk3012@columbia.edu}
\author[4]{\fnm{Ruoxi} \sur{Jia}}\email{ruoxijia@vt.edu}
\author[2]{\fnm{Carole-Jean} \sur{Wu}}\email{carolejeanwu@meta.com}
\author[5]{\fnm{Matei} \sur{Zaharia}}\email{matei@berkeley.edu}
\author[1]{\fnm{James} \sur{Zou}}\email{jamesz@stanford.edu}

\affil[1]{
\orgname{Stanford University}, \orgaddress{ \city{Stanford}, \state{CA}, \country{USA}}}

\affil[2]{\orgname{FAIR, Meta}, \orgaddress{ \city{Menlo Park}, \state{CA}, \country{USA}}}


\affil[3]{
\orgname{Columbia University}, \orgaddress{ \city{New York}, \state{NY}, \country{USA}}}


\affil[4]{
\orgname{Virginia Tech}, \orgaddress{ \city{Blacksburg}, \state{VA}, \country{USA}}}

\affil[5]{
\orgname{University of California, Berkeley}, \orgaddress{ \city{Berkeley}, \state{CA}, \country{USA}}}

\abstract{As Machine Learning (ML) systems continue to grow, the demand for relevant and comprehensive datasets becomes imperative.
There is limited study on the challenges of data acquisition due to ad-hoc processes and lack of consistent methodologies.
We first present an investigation of current data marketplaces, revealing lack of platforms offering detailed information about datasets, transparent pricing, standardized data formats.
With the objective of inciting participation from the data-centric AI community, we then introduce the \textit{DAM challenge}, a benchmark to model the interaction between the data providers and acquirers in a data marketplace. The benchmark was released as a part of DataPerf~\cite{mazumder2022dataperf}. Our evaluation of the submitted strategies underlines the need for effective data acquisition strategies in ML. 
}



\maketitle
\section{Introduction}\label{sec1}
\input{1_intro}

\section{Overview of Existing Data Marketplaces for ML}
\input{2_overview}

\section{Data Acquisition for ML Benchmark: DAM}
\input{3_benchmark}

\section{Results}
\input{4_evaluation}

\section{Looking Forward}
\input{5_future}

\section{Conclusion}
\input{6_conclusion}

\section{Acknowledgements}
Thank you to Ce Zhang, Mostafa Elhoushi, Luis Oala, Max Huang, Sudnya Diamos, Danilo Brajovic, Hugh Leather and the DataPerf organizers who gave feedback in designing the data acquisition challenge and during alpha testing.
Thanks to Rafael Mosquera for his feedback on the benchmark and his efforts in supporting DAM in DynaBench.


\bibliography{sn-bibliography}
\newpage
\appendix
\input{_appendix}

\end{document}

%% file: 1_intro.tex
Datasets, the cornerstone of modern machine learning (ML) systems, have been increasingly sold and purchased for different ML pipelines~\cite{pei2020survey}. 
Several data marketplaces have emerged to serve different stages of building ML-enhanced data applications. For example,  NASDAQ Data Link~\cite{nasdaq} offers financial datasets cleaned and structured for model training, Amazon AWS data exchange~\cite{aws_data_exchange} focuses on generic tabular datasets,  and Databricks Marketplace~\cite{databricks_marketplace} integrates raw datasets and ML pipelines to deliver insights. 
The data-as-a-service market size was more than 30 billions and is expected to double in the next five years~\cite{DaaS_MarketAnalysis}. 
 
While the data marketplaces are increasingly expanding, unfortunately, data acquisition for ML remains challenging, partially due to its \textit{ad-hoc} nature:
Based on discussions with real-world users, data acquirers often need to negotiate varying contracts with different data providers first, then purchase multiple datasets with different formats, and finally filtering out unnecessary data from the purchased datasets.
This is inefficient since negotiation requires tremendous human efforts, while purchasing datasets which are later filtered out leads to a waste of money.

Information opaqueness and lack of principles are the main factors for such an inefficiency.  Most data providers are reluctant to offer the full details of their datasets to data acquirers.
Consequently, it is challenging for the data acquirers to design principled data acquisition strategies. 
This is potentially a lose-lose: acquirers fail to identify the desired datasets for their applications, while data providers abandon a large fraction of users and thus lose their revenues. 
Thus we ask: \textit{how can we design a data marketplace for ML which offers budget-awareness, information and price transparency, and multiple data sources?}

Addressing these important challenges  requires not only individual researchers or companies but  collaborative efforts from the entire data-centric AI community. 
To encourage community efforts, we give an in-depth analysis of the existing data marketplaces, and identify three important desiderata of a data marketplace: (i) transparent pricing, (ii) unified data format, and (iii) ML-aware acquisition guidance.
Thus, we design the DAM challenge, a benchmark for a data marketplace that offers all the desiderata and solicits ML-aware data acquisition strategies. As part of the MLCommons DataPerf initiative~\cite{mazumder2022dataperf}, the first launch  has attracted promising solutions.
Our discussion and analysis of the received strategies underscore the importance of developing data acquisition strategies, as large performance gaps between different strategies exist, and no single strategy outperforms others for all data market instances we consider. 
Overall, we hope this paper lays a foundation for data acquisition in data-centric AI and stimulates a broad range of researchers to tackle important challenges in the area.

\eat{As the cornerstone of machine learning (ML) systems, datasets are increasingly sold and purchased for different ML pipelines. Several data marketplaces have emerged such as Amazon AWS data exchange~\cite{aws_data_exchange} and NASDAQ Data Link~\cite{nasdaq}, and the data-as-a-service market size was more than 30 billions and is expected to double in the next five years~\cite{DaaS_MarketAnalysis}. While increasingly expanding, new challenges arise in the data marketplace for both data providers and acquirers.

One of the main challenges in data marketplaces arise from limited information share between data providers and acquirers prior to data set acquisition. Data content is usually opaque: data providers usually are disinclined to release the full content of their datasets to the acquirers. From acquirers' perspective, this makes it challenging to decide whether a dataset is useful and high quality for the downstream ML tasks. High quality data is crucial in building trustworthy AI applications~\cite{liang2022advances} and it is difficult to ensure the quality with limited information. Datasets often require substantial work in selection, cleaning and annotation.

In this paper, we provide an analysis of existing major data marketplaces for ML, identify the challenges in the market, and design a benchmark to evaluate different algorithmic solutions towards those challenges.
This benchmark has been released as part of the MLCommons DataPerf initiative~\cite{mazumder2022dataperf}.}

\begin{figure*}[t]
	\centering
	\vspace{-0mm}
	\includegraphics[width=0.999\linewidth]{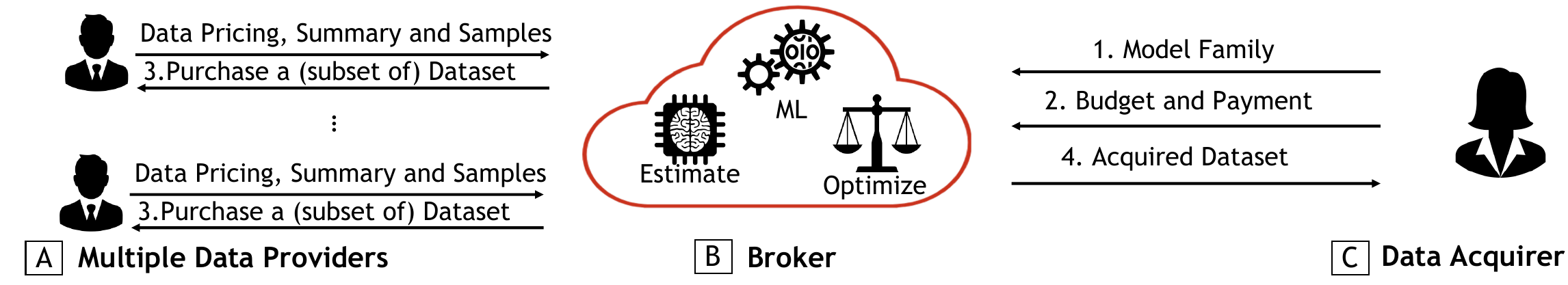}
	\vspace{-0mm}
	\caption{Overview of the data acquisition for machine learning marketplace. It consists of three agents: data providers, a broker, and a data acquirer. The data providers publicly release their  pricing mechanisms, data summaries, and a few samples from their datasets.  The data acquirer first gives the broker (i) the model family she is interested in training on the purchased data samples, (ii) her own evaluation data, and (iii) the budget she is willing to spend as well as the  payment. Next, the broker decides which datasets to purchase as the training data to optimize the model performance on acquirer's data. Finally, it acquires corresponding datasets from the providers and send it back to the acquirer. The DAM benchmark simulates both providers and the acquirer, and ask the participators to construct a broker as good as possible.}
	\label{fig:DAM:IntroOverview}
\end{figure*}

%% file: 2_overview.tex
\subsection{What type of data acquisition services are there?}

\begin{figure*}[t]
	\centering
	\vspace{-0mm}
	\includegraphics[width=0.90\linewidth]{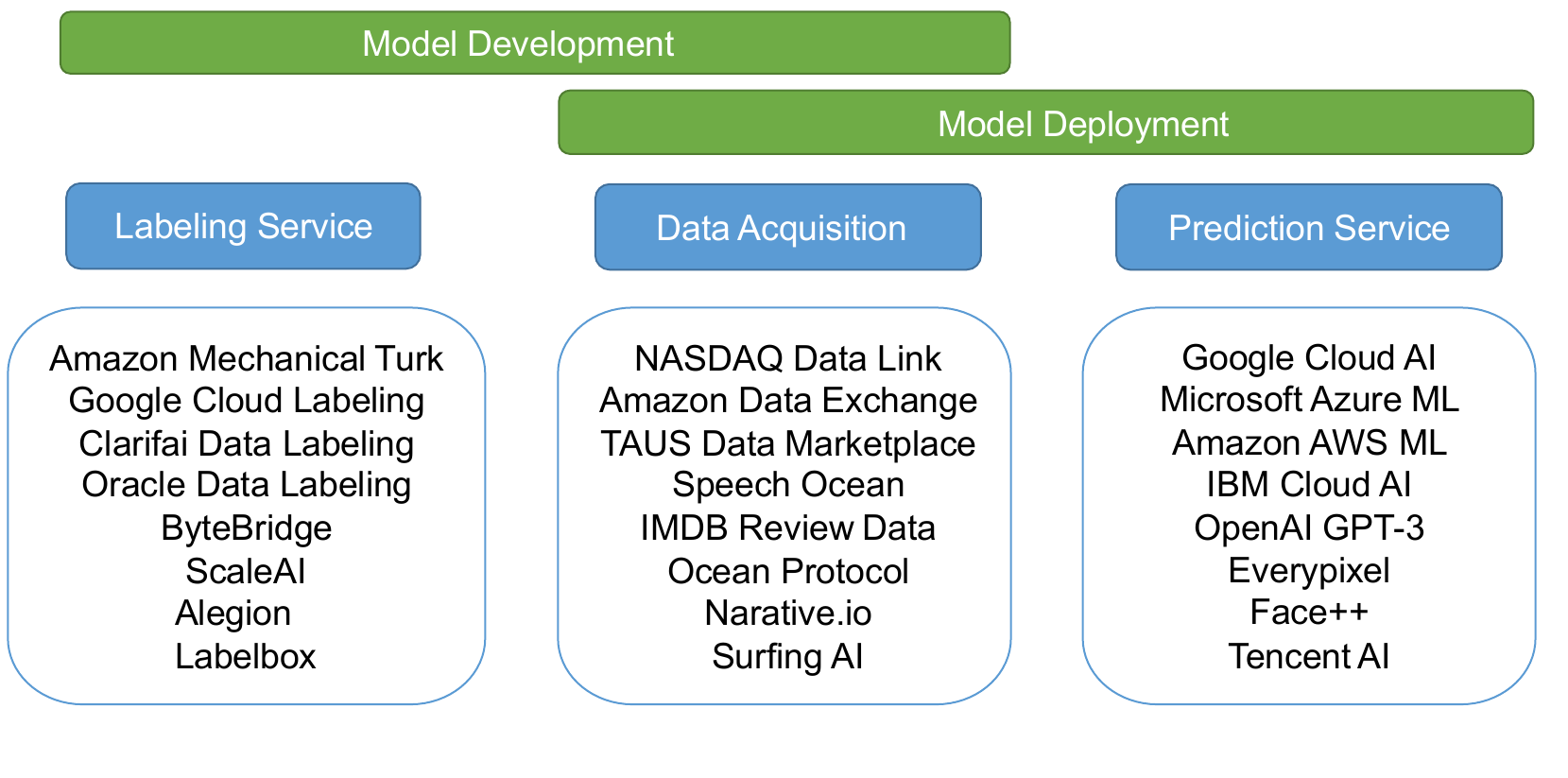}
	\vspace{-2mm}
	\caption{Data Service Types. (i) The long-standing labeling services offer annotations to data (such as images, texts, and audio) provided by the customers. (ii) On the other hand, data acquisition services take the users' description as input, and then returns desired data with or without annotations. (iii) Prediction service emerges as a new data service: it produces machine generations on any given inputs.}
	\label{fig:DAM:DataServiceTypes}
\end{figure*}

Data marketplace for ML is broad and has various forms of commodity that is sold and purchased (see Figure \ref{fig:DAM:DataServiceTypes}). These include labeling services, data acquisition in the model development stage and prediction services in the model deployment stage. Here, the queries are generic and include (i) human labeling services on the dataset, (ii) raw data acquisition, (iii) some data products (such as an ML service) built on top of it. Most data providers adopt (i) and (ii). More recently, more data providers are selling data products (iii) such as ML services. For example, Google uses their own datasets to build vision services, i.e. Google Vision API, which give annotations to user data for a fee~\cite{google_vision_api}.     
While all of the mentioned data services are important, we focus data markets for raw data in this work.

\subsection{Why is raw data acquisition needed for training ML models?}

A natural question is why data acquisition is needed given the abundant amount of publicly available data, such as ImageNet~\cite{imagenet} consisting of millions of natural images, SQuAD 2.0 containing more than one million English question-answer pairs~\cite{rajpurkar2018know}, Common Crawl including petabytes of webpage text data~\cite{commoncrawl}. 
For many downstream tasks, however, publicly available datasets lack the diversity needed to represent real-world scenarios and frequently suffer from quality issues.

For instance, in the case of Chinese speech recognition, publicly available utterances are mostly recorded in quiet environments, which do not accurately reflect real-world scenarios with diverse noises and delays. Moreover, the speakers in these utterances primarily use standard mandarin, whereas different dialects exhibit distinct pronunciations of the same words or phrases, and some even contain slangs that do not exist in standard mandarin. In the absence of training data that covers these missing contexts, achieving decent performance during inference can be challenging.

Even when publicly available training data covers all possible contexts and domains, the quality of the data remains a concern. Annotation errors are prevalent in many open datasets, such as ImageNet, which can significantly limit the performance of any machine learning models trained on them. In contrast, training on high-quality datasets purchased from professional companies can generate a much higher upper bound on achievable performance.

\subsection{How does data acquisition for ML happen?}
A data marketplace for ML is captured by participants, data or data services, interactions, pricing, and contracts. Participants include data providers, who want to sell their data, data acquirers, who need to acquire data for their own ML applications, and sometimes data brokers, who serve as a middleman between data providers and data acquirers. 

For any downstream task, there are often several potential data providers.
Data can be sold in bulk as curated datasets or as individual data points.
Each provider gives a description of its own dataset, a pricing mechanism, and potentially a few samples from the dataset. 
There is often some usage term of use associated with the dataset. 
The most important restriction is that the dataset cannot be further sold by the acquirer.
This is due to the licensing restrictions.

Table~\ref{fig:marketplaces} shows a list of some of the existing data marketplaces and categorizes their domain, interaction model, transaction type and pricing model. The main takeaway is that the market is ad-hoc. Different types of data are sold and purchased from different domains. In terms of information shared before purchase, most common practice is to share only metadata information about the dataset or a few data samples. Transaction type also varies: some marketplaces has one time upfront payment, some are subscription based and some charges based on API usage. The prices are sometimes public but in the majority of the markets prices are not advertised publicly and contacting sales is required.
Here we expand on the properties in data marketplaces and list the challenges we observe in current marketplaces.

\textbf{Roles.} Data provider, data acquirer and broker are the three main roles in the marketplace. A broker is not always necessary -- some of the data providers offer their data directly to the acquirers without a third party broker, such as Twitter API~\cite{twitter_api}, Nasdaq Data Link~\cite{nasdaq}. On the other hand, brokers can make data access and management
easier, especially if tied with a compute platform. For example Amazon AWS Data Exchange~\cite{aws_data_exchange} and Databricks Marketplace~\cite{databricks_marketplace} offer access to a variety of data providers' data to customers through their platform. From acquirers' perspective, having a single platform where multiple data providers data can be found makes it easier to search and find the relevant data. However in the current marketplaces, there are variety of data providers that do not offer their data through a broker platform. For acquirers this makes access to data harder  due to dis-aggregation and for providers it might make it harder to reach to customers.

\textbf{Domains.} There are various domains in the data marketplaces such as vision, speech, NLP, finance, healthcare, etc. Some of the marketplaces are not focused on a particular domain; for example AWS Data Exchange~\cite{aws_data_exchange} and Databricks Marketplace~\cite{databricks_marketplace} includes data providers from a broad range of domains.
On the other hand some of the brokers are focused on one specific domain, such as Gradient Health~\cite{gradient} and Narrative~\cite{narrative}.
Gradient Health is focused on medical imaging data and gathers patient data from various hospitals. Narrative is focused on demographic and location data gathered from different data providers.
Focusing on a particular domain allows these platforms to offer custom features specific for their data type, such as allowing data acquirers to select different attributes from the data and filter it before they make the purchase. For example; Gradient Health allows filtering data by imaging type. Narrative allows filtering people data by age and location.
Due to the domain specific nature, each domains would require a different set of attributes that cannot be generalized.

\textbf{Interaction Types.}
Interaction between the providers and acquirers before making a purchase is critical. The acquirers need information about the dataset properties to validate whether the dataset is useful for their applications. 
However providers often are not willing to share their dataset prior to purchase and acquirers are not willing to share their use case or models due to confidentiality. This creates the biggest challenge in the marketplace, how to evaluate the value of the data with limited information?

Most of the existing research assumes the providers or acquirers are willing to share their full data~\cite{agarwal2019marketplace} or significant number of data samples~\cite{kang2023performance}, however in current marketplaces
the information shared prior to the purchase of the data is extremely limited.
Typical interaction includes data providers to share (i) a few samples from their datasets, (ii) certain meta-data, (iii) summary statistics on the dataset.
For example TAUS, Magic Data, Datatang, Core Signal are examples of data provider
that share only a few samples from the datasets. AWS Data Exchange, Databricks Marketpace and Speech Ocean provides only some metadata and description of the datasets without any samples.

\textbf{Transaction Models.}
Popular transaction methods include (i) one-time upfront pricing,  (ii) query-based pricing, and (iii) subscription pricing. One-time pricing assigns a fixed price for any given dataset. This works well if the dataset is fixed and relatively small. Query-based pricing allows for sharing a small part of the dataset. For example, one can get 5\% of the entire dataset, and only pay for a small amount of dollars. This works when the entire dataset is too large and acquirers cannot afford buying the whole. Subscription pricing gives the users the dataset access only for a fixed period of time.

\textbf{Pricing.} This aspect considers whether a data has a fixed price that is visible to all potential data acquirers or has a negotiable price that is not visible publicly. The majority of the marketplaces falls into the second category and they do not show the price publicly. 
During private price negotiations, providers may offer less price per data sample if acquirer purchases in bulk / more data samples.

\textbf{Data Format.}
Data can be sold as curated datasets in bulk or as individual data points/samples. Some marketplaces do allow filtering of data based on some features or criteria, however price may not increase linearly with each data point to purchase and buying as bulk can often be more cost effective.
Another major challenge in the data marketplaces is the varying data file formats.
For a data acquirer, this makes combining data from multiple sources challenging since it requires additional work to convert data formats from different providers into a common format.
To address this problem, there are efforts in the industry to unify the data format for ML training, such as Croissant~\cite{croissant}. Croissant is a \textit{high-level format for machine learning datasets that combines metadata, resource file descriptions, data structure, and default ML semantics into a single file}.

\begin{table}[]
\scalebox{0.7}{
\begin{tabular}{l|c|c|c|c|c}
\hline
\multicolumn{1}{c|}{\textbf{Data Market}} & \multicolumn{1}{c|}{\textbf{Role}} &  \multicolumn{1}{c|}{\textbf{Domain}} & \multicolumn{1}{c|}{\textbf{Interaction}} & \multicolumn{1}{c|}{\textbf{Transaction}} & \multicolumn{1}{c}{\textbf{Pricing}} \\ 
\multicolumn{1}{c|}{} &  \multicolumn{1}{c|}{} & \multicolumn{1}{c|}{\textbf{Model}} & \multicolumn{1}{c|}{\textbf{Type}} & \multicolumn{1}{c|}{\textbf{Model}} & \multicolumn{1}{c}{\textbf{Transparency}} \\ \specialrule{.12em}{.05em}{.05em} 

AWS Data Exchange~\cite{aws_data_exchange} & Broker & Varying & Metadata & Upfront/Subscription & Hidden  \\ \hline
Databricks Marketplace~\cite{databricks_marketplace} & Broker & Varying & Metadata  & Unknown & Hidden  \\ \hline
Narrative~\cite{narrative} & Broker & Varying & Metadata & Query based & Hidden  \\ \hline
TAUS~\cite{taus} & Broker & NLP/Translation & A few samples & Upfront & Fixed   \\ \hline
PromptBase~\cite{prompt_marketplace} & Broker & Prompts for GenAI & Sample output & Upfront & Fixed  \\ \hline
Gradient Health~\cite{gradient} & Broker & Healthcare & Metadata & Query based & Hidden \\ \hline
Snowflake~\cite{snowflake_datamarket} & Broker & Varying & Metadata & Query based & Fixed \\ \hline 
Speech Ocean~\cite{speechocean} & Data Provider & Speech, Vision & Metadata & Unknown & Hidden  \\ \hline
Magic Data~\cite{magicdata} & Data Provider & Speech, Vision, NLP & A few samples & Unknown & Hidden \\ \hline
Datatang~\cite{datatang} & Data Provider & Speech, Vision, NLP & A few samples & Unknown & Hidden  \\ \hline
Surfing Tech~\cite{surfingtech} & Data Provider & Speech, Vision & Unknown & Unknown & Hidden \\ \hline
Core Signal~\cite{coresignal} & Data Provider & Business, Recruitment & A few samples & PAYG & Hidden  \\ \hline
NASDAQ Data Link~\cite{nasdaq} & Data Provider & Finance  & A few samples & Subscription & Fixed  \\ \hline
Twitter API~\cite{twitter_api} & Data Provider & Social Media & A few samples & Subscription & Fixed  \\ \hline
\end{tabular}
}
\vspace{0.5mm}
\caption{Examples of data marketplaces and their features. These data marketplaces offer differ in who provides the data, which domain their data comes from, how potential buyers interact with them, the pricing model and transparency.}
\label{fig:marketplaces}
\end{table}

\subsection{Challenges and opportunities in data marketplaces}
Data marketplaces present several challenges, such as ensuring data quality, addressing privacy and security concerns, and creating a fair and transparent pricing system. However, these challenges also present opportunities for innovation. An ideal data marketplace would have several key properties, as shown in Table~\ref{fig:properties}. Firstly, it would have \textit{budget awareness}, where data acquirers can easily understand the cost of the data they are purchasing and make informed decisions about their budget. Secondly, it would have \textit{price transparency}, where data providers can openly communicate their pricing models and data acquirers can compare prices across different providers. Thirdly, it would have \textit{multiple data providers}, offering a diverse range of data sources and allowing data acquirers to choose the best data for their needs. Finally, it would have \textit{useful information sharing}, where data acquirers and data providers can share information and insights to improve the quality and relevance of the data being sold. Yet, none of the existing data marketplaces satisfy all four properties. 

In such an ideal marketplace, data acquirers would have access to a wide range of high-quality data from multiple providers, allowing them to make more informed decisions and drive better business outcomes. Data providers, on the other hand, would have a platform to showcase their data and compete on price and quality, leading to increased competition and innovation. Additionally, the marketplace could offer features such as data validation and cleaning, ensuring that the data being sold is accurate and reliable. Overall, an ideal data marketplace would provide a transparent, competitive, and innovative environment for buying and selling data, ultimately benefiting both data providers and acquirers.
With this goal, we designed DAM, Data Acquisition Benchmark for ML (DAM), which we explain next.

\begin{table}[hb]
\scalebox{1}{
\begin{tabular}{l|c|c|c|c}
\hline
\multicolumn{1}{c|}{\textbf{Properties}} & \multicolumn{1}{c|}{\textbf{DAM}} &  \multicolumn{1}{c|}{\textbf{AWS Data}} & \multicolumn{1}{c|}{\textbf{Taus\cite{taus}}} & \multicolumn{1}{c}{\textbf{Projector\cite{kang2023performance}}} \\
\multicolumn{1}{c|}{\textbf{}} & \multicolumn{1}{c|}{\textbf{}} &  \multicolumn{1}{c|}{\textbf{Exchange\cite{aws_data_exchange}}} & \multicolumn{1}{c|}{\textbf{}} 
\\ \specialrule{.12em}{.05em}{.05em}

Budget Awareness & \cmark & \xmark & \xmark & \cmark \\ \hline
Price Transparency & \cmark & \xmark & \cmark & \cmark \\ \hline
Useful Information Share & \cmark & \xmark & \xmark & \xmark \\ \hline
Multi-Provider Support & \cmark & \cmark & \cmark & \cmark \\ \hline
\end{tabular}
}
\vspace{3mm}
\caption{Properties of existing mainstream data marketplaces. AWS Data Exchange supports multiple data providers, but their pricing mechanism is often opaque. AWS Data Exchange and Taus gives no budget control. All existing data marketplaces lack a systematic way to share useful information with the potential buyers before transactions. To the best of our knowledge, DAM is the first benchmark for a data marketplace for ML that satisfies all desiderata. } 
\label{fig:properties}
\end{table}

%% file: 3_benchmark.tex

Based on our observations and challenges in the current data marketplaces, we designed a benchmark, Data Acquisition for ML (DAM), with the goal of mitigating a data acquirer’s burden by automating and optimizing the data acquisition strategies. In this section, we provide the overall design of DAM along with a concrete instantiation.

\subsection{Market Setups and Problem Statement}
In DAM, we consider a data marketplace consisting of $K$ data providers and one data acquirer. 
Each provider $i$ holds a labeled dataset to sell, denoted by $D_i$.  Note that $\|D_i\|$, the size of these datasets, can vary. 
To encourage acquirers with varying affordability, data providers allow purchasing subsets of their datasets.
For example, one may purchase the entire dataset $D_i$, or only 25\% or 50\% data points from $D_i$.
The price then naturally depends on the number of the purchased samples.  
Formally, 
we denote the pricing function for $D_i$  by $p_i: \mathbb{N} \to \mathbb{R}{^+}$.
If $q \in \mathbb{N}$ samples from $D_i$ is purchased, then one needs to pay $p_i(q)$.  
The pricing function is non-negative and monotone with respect to the number of samples.

\paragraph{What pre-acquisition information to share with the buyer?} Demonstrations play an essential role in both traditional and data markets. 
In traditional markets, directly exhibiting the product is a natural way to attract potential buyers. 
Our discussion with real-world data providers indicates, however, that revealing considerable data instances before the acquirer decides to buy anything is not desired, as the value of the datasets can be lost due to data revealing.
Thus, DAM only requires providers to reveal only a small amount (=5) of samples. In addition, summary statistics that describe high-level features of datasets are often showcased by existing data marketplaces~\cite{taus,narrative} to attract potential buyers. Thus, DAM also reveals summary statistics on the datasets.

More formally, we use $\mathcal{L}_i$ and $s_i$ to denote the list of shared samples and the summary statistics for the $i$th provider. 
The data acquirer observes the list of shared samples, the summary statistics and pricing functions, $\{(\mathcal{L}_i, s_i, p_i(\cdot))\}_{i=1}^{K}$. 
A budget $b \in \mathbb{N}$, a small evaluation dataset $D_b$, and a training model $f(\cdot)$. The distribution of the evaluation dataset is not necessarily the same as the datasets sold by the data providers. In fact, part of the key challenge of data acquisition is to find which data is ``similar'' enough with the evaluation data before buying it.
The acquirer's goal is to identify a purchase strategy $(q_1, q_2,\cdots, q_K) \in \mathbb{N}^{K}$ and $0 \leq q_i \leq  \|D_i\|$ for all $i$, such that the total cost is within the budget $b$, and the accuracy of the ML model $f(\cdot)$ on the evaluation dataset $D_b$ is maximized when it is trained on the purchased datasets. 
The details of the  summary statistics as well as the pricing functions will be given in the next subsection.

\subsection{Sentiment Analysis on Different Data Providers: A Concrete Instantiation}
Here we consider a concrete instance of the above design. 

\textbf{Setup of the marketplace.} We consider $K=20$ different data providers. Each of them is selling a dataset for sentiment analysis. Each data point in a dataset $D_i$ is a pair of (i) a feature vector representing the embedding of some text paragraphs, and (ii) a label indicating the nuance of an opinion (e.g., positive or negative) in the text. 
All providers use the same feature extractors to encode their raw datasets. The  quality of data labels also varies across different data providers. The specific data preprocessing details shall be released after the competition is retired.  
To overcome potential overfitting, we have created five distinct market instances. The structure of each market is identical: 20 data providers, 1 buyer, and the same type of information to share. 
On the other hand, the data points sold by each provider are sampled from a large-scale data pool using different sampling distributions.The original data pool contains 21 categories. 
For each data provider, we sample different number of samples from each category. 
The different samples simulate a diverse marketplace. 
Each marketplace is also unique due to the varying number of samples from each category.

\textbf{ Summary statistics}. 
In our instantiation, the summary statistics contain (i) the 100-quantiles of the marginal distribution  of each feature as well as the label and (ii) the correlations between each feature and the label. 
These summary statistics were selected to offer useful insights on the provider's data while keeping their data secure and private. 

\textbf{Pricing functions.} Each dataset is worthy \$100 and a linear pricing function is adopted. 
Note that the number of samples within each dataset is not necessarily the same.

\textbf{Acquirers' Tasks.} The acquirer holds a small dataset with the same structure (embedding vectors and labels). 
A logistic regression model is used as the ML model.
The acquirer's budget is \$150. Each submitter's goal is to figure out the purchase strategy $(q_1, \cdots, q_K)$.
After this, each fraction $q_i$ can be converted to the number of samples to purchase via $q_i$ to obtain the number of samples to purchase from each provider.

\textbf{Evaluation.} How to quantify the performance of a strategy? For each market instance, we first compute the following score (normalized by 100):

$$\mathrm{score} \triangleq 100 \cdot \left( \alpha \times \mathrm{Accuracy} + (1-\alpha) \times \frac{\mathrm{budget} - \mathrm{cost}}{\mathrm{budget}}  \right) $$
Then we use the average of the five market instances as the final metric. Here, the goal is to maximize the overall accuracy while minimizing the cost. The factor $\alpha$ controls how much budget saving is appreciated. In the existing version of the DAM benchmark, we set $\alpha=0.98$, encouraging submitters to focus primarily on accuracy. 

\subsection{Solutions}
Here, we present the solutions submitted by the benchmark participants.

\paragraph{Strategy-Single:} The first strategy is to purchase a single provider's data points as many as possible within the budget $b$. To be more specific, this strategy first selects a provider $i \in [K] := \{1, \dots, K\}$ and purchase $\min(\|D_i\|, n_i)$ data from the $i$-th provider where
\begin{align*}
    n_i = \mathrm{argmax}_{x \in \mathbb{N}} p_i(x) \leq b.
\end{align*}
In DAM, the total price of each provider's dataset is always less than the budget, \textit{i.e.}, $p_i(\|D_i\|) \leq b$, resulting in buying the entire dataset. After the purchase, the remaining budget is exactly one-third of the total budget. We denote this strategy by \textsf{Strategy-Single-$i$} where $i$ indicates the selected provider's identifier. 

\paragraph{Strategy-All:}
The second strategy is to purchase data from every provider with an equal amount of budget for each provider. In contrast to \textsf{Strategy-Single-$i$}, this approach allows us to spend the entire budget, and it is no longer required to select a specific provider. This strategy is expressed as follows. For all $i \in [K]$, we bought $n_i$ data from the provider $i$ where 
\begin{align*}
    n_i = \mathrm{argmax}_{x \in \mathbb{N}} p_i(x) \leq \frac{b}{K}.
\end{align*}
We denote this strategy by \textsf{Strategy-All}.

\begin{figure}[t]
  \centering
    \centering
    \vfill
    \includegraphics[width=1\textwidth]{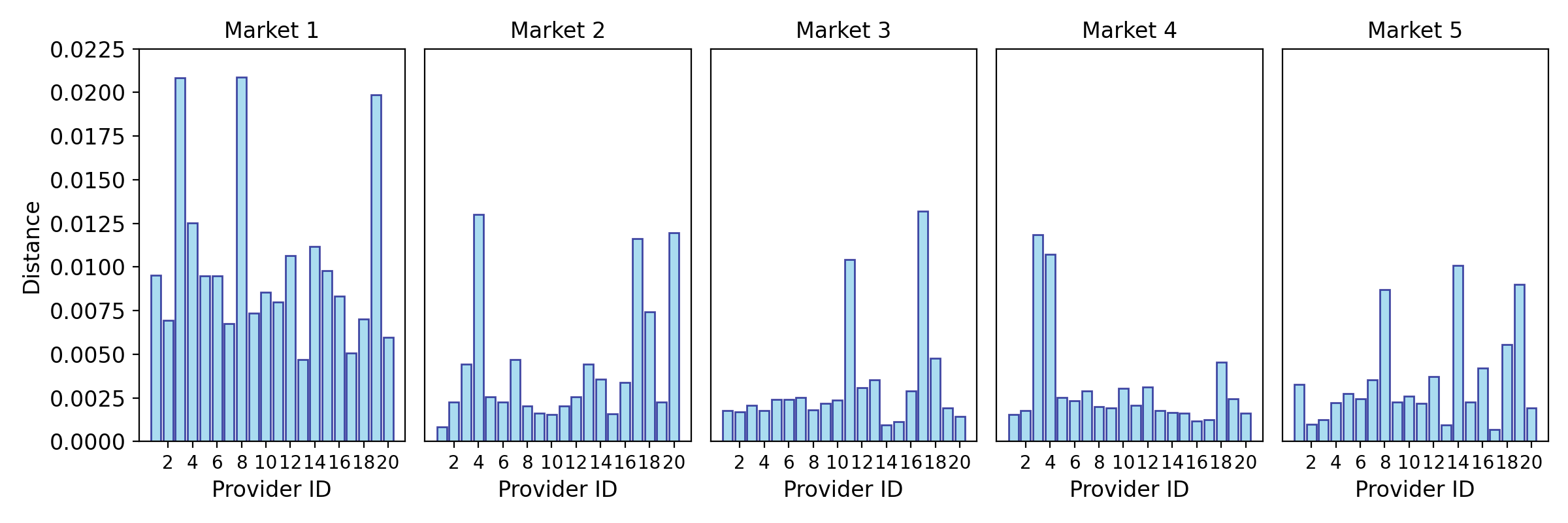}
    \caption{
    The Euclidean distance of label correlations between the acquirer and the provider data across the five markets. This calculation helps identify providers whose label correlations significantly differ from those of the acquirer. Larger distance values indicate greater dissimilarity. For instance, in Market 1, three providers (ID \#3, \#8, and \#19) exhibit notably high distance values, indicating substantial dissimilarity from the acquirer.}
    \vfill
  \label{fig:combined_label}
\end{figure}

\paragraph{Strategy-$p\%$}
Our third strategy is to purchase data from a subset of data providers by leveraging the distributional similarity between the acquirer and providers. To be more specific, we denote the correlation coefficients between the label and the $k$-th feature within the acquirer dataset by $r_{\mathrm{acquirer}} ^{(k)} \in \mathbb{R}$ and set $r_{\mathrm{acquirer}} :=(r_{\mathrm{acquirer}}^{(1)}, \dots, r_{\mathrm{acquirer}} ^{(d)})$ where $d$ is the input dimension. Analogously, for $j \in [K]$, a vector $r_{\mathrm{provider}, j} \in \mathbb{R}^d$ denotes correlation coefficients between the label and each feature within the $j$-th provider dataset. We then calculate the Euclidean distance between acquirer and provider vectors.
\begin{align*}
    Q_{j} := \norm{r_{\mathrm{acquirer}} - r_{\mathrm{provider},j}}_2 ^2 
\end{align*}
Figure~\ref{fig:combined_label} illustrates the distribution of $Q_j$ across the five different markets. It shows there are several providers whose label correlations are more different from those of the acquirer than others. Based on this observation, we exclude $p\%$ of providers whose $Q_j$ values are larger than others and apply \textsf{Strategy-All} to the remaining providers. We call this strategy \textsf{Strategy-$p\%$}.

\paragraph{Strategy-RFE (\underline{R}ecursive \underline{F}eature \underline{E}limination)}

Due to the higher dimensional nature of the data (768) and not knowing any of its structure, we reduce the input dimensionality through standard \textit{feature selection with recursive feature elimination (RFE) \cite{darst2018using}}.
Specifically, this backward elimination procedure starts with training the target model with all features. At each time, it removes the feature with the weakest impact on the model's prediction and re-train the model with the remaining features.
In our case, the strength of each feature is measured by the corresponding model coefficient's absolute value. This process iterates until the target number of features is reached. The remaining set of features is considered to be most essential to the model's prediction. This helps to find the most important features and refine our analysis to the reduced data.

For each provider's data, the correlation score between each feature to the prediction variable is provided. For ease of elaboration, we refer to this correlation score as \textit{feature relevance} hereafter. Our hypothesis is that if a provider's data is consistent with the acquirer's data and works similarly with target the model, we should observe a high consistency between coefficients of the model trained on the acquirer's data and the feature relevance of the provider's data. For example, for a given feature, if the coefficient of the trained model is positive, which indicates that an increase in this feature increases the chance for the model to predict a positive label, we would expect the correlation between the value of this feature and the label (i.e., feature relevance) to also be positive.

\begin{figure}[t]
  \centering
    \centering
    \vfill
    \includegraphics[width=1\textwidth]{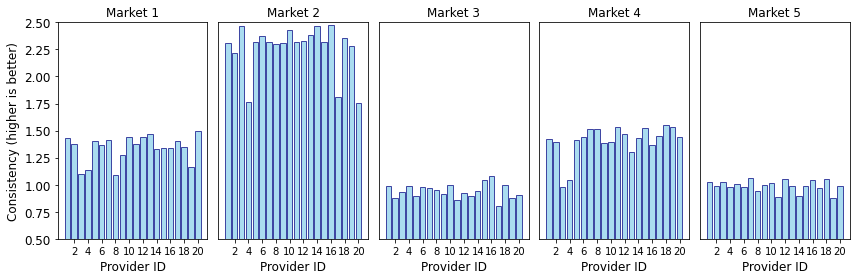}
    \caption{Consistency (measured as dot product on normalized vectors) between acquirer's data and data from each provider evaluated on features selected via RFE. Some data markets demonstrate high divergence compared to others. In general, we found this consistency measure can effectively identify low-quality data providers (which are marked by a much lower consistency in this measure). Nonetheless, its ability to distinguish higher-quality data providers is less remarkable. Many data providers achieve similar scores on this consistency measure and it is hard to further differentiate them by this score alone.}
    \vfill
  \label{fig:sol1}
\end{figure}

Thus, for features selected by RFE, we first train a logistic regression model on the acquirer's data to obtain the coefficients. A high value on this measure should imply that the data from a provider is more consistent with the validation data such that it better suits the task. Results are visualized in Figure \ref{fig:sol1}. We normalize both coefficients and feature relevance to between 0 and 1 and calculate the \textit{dot product} between the two as the similarity measure. We select the highest valued two datasets, where we select the maximum possible samples for the top 1 dataset and allocate any remaining budget to the second runner-up.  Note that this scheme does not take account into the effect of different costs for the data. So we skip the data providers with a higher data cost per sample than the others, which are provider 8 for data markets 2, 3, 4 and provider 9 for data markets 3, 4, 5, respectively.

\paragraph{Strategy-CoFR (\underline{Co}sine similarity importance-\underline{F}eature \underline{R}elevance)}

As opposed to Strategy-FRE, which examines
the top 5 important features selected by RFE, \textit{in this strategy, we calculate the consistency measure (normalized dot product) across all 768 features.} As consistency measure is essentially a proxy to \textit{cosine similarity}, we refer to this strategy as \textit{CoFR (\underline{Co}sine similarity importance measure-\underline{F}eature \underline{R}elevance)}. Results are shown in Figure~\ref{fig:sol2} in Appendix. Same as in Strategy-RFE, we select the top two data providers with the highest correlation to acquirer's data–selecting maximum samples from the first provider, allocating the remaining budget to the second runner-up, and avoiding high-cost data providers.

\paragraph{Strategy-$L_P$}
Similar to Strategy-CoFR, in this strategy, we calculate the \textit{$L_P$ distance between normalized coefficients of the model trained on acquirer's data and the feature relevance for data from each provider} on all 768 features, where a small distance implies high consistency. We examine $L_2$, $L_1$, and $L_\infty$ distances, respectively. The results are depicted in Figure~\ref{fig:sol31} in Appendix, respectively. Selection scheme is the same as in Strategy-RFE and Strategy-CoFR. Selections for $L_2$ and $L_1$ distances ended up exactly identical.

%% file: 4_evaluation.tex
\begin{figure}[t]
    \centering
    \includegraphics[width=1.05\textwidth]{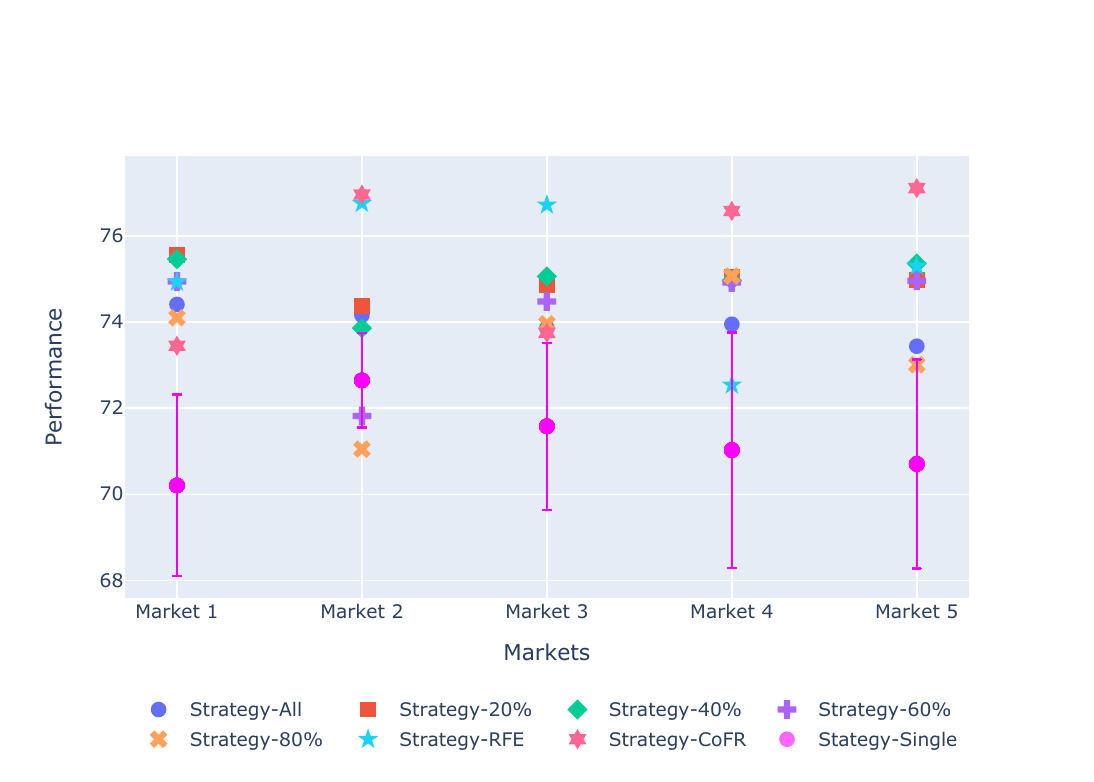}
    \caption{Evaluation of the Proposed Solutions. The pink point denotes the average performance when randomly selecting $i$ and then adopt strategy-single-i, while its error bar indicates one quarter of the standard deviation. Strategy-$\ell_p$is removed for robust visualization, and its performance can be found in Table \ref{tab:performance}.  We observe that no strategy outperforms all others universally. For example, the CoFR approach ranks the first on the second, fourth, and fifth market instance. However, RFE is better for the third market, and Strategy-20\% and Strategy-40\% rank the top-2 positions for the first market. The variance of Strategy-Single is large. If picking the right single provider, it may achieve the highest performance, which in practice, however, is challenging to do before purchase. } 
    \label{fig:benchmarkresult}
\end{figure}

We have evaluated all proposed strategies on the five distinct data marketplaces. The results are presented in Figure~\ref{fig:benchmarkresult}, and the details can be found in Table \ref{tab:performance} in the Appendix.

There are several interesting observations. First, there is no universally ``best'' strategy. 
For example, the Strategy-CoFR approach gives the best performance on the second, fourth and fifth data marketplace. However,
Strategy-RFE is better for the third market, and Strategy-20\% and Strategy-40\% are the top-2 for the first marketplace. This underscores the importance of carefully customizing the data acquisition strategies for different marketplaces.
Second, there is a large variance for the Strategy-Single approach. In fact, we observe that Strategy-Single-20 is often the best strategy, and  Strategy-Single-3 and Strategy-Single-8 lead to limited performance. Detailed list of results are shown in Table~\ref{tab:performance} in Appendix. 
In practice, however, it is challenging to predict which data provider leads to the best or worst performance, and randomly picking one ends up with limited performance.  

%% file: 5_future.tex
\subsection{Alternative Data Acquisition Benchmark Designs}

\eat{\TODO{@Feiyang: add the reference and discussion on what info to  share part.}
}



There can be various alternative benchmark designs that are useful for data marketplaces. In this section, we discuss some of the useful scenarios we identified.

\textbf{Pre-acquisition Evaluation:} 
Given the limited information provided by data providers, an important challenge faced by an acquirer is to estimate how well a model trained on the provider’s data performs on the acquirer's data seen during deployment (in terms of accuracy, f1, mAP, etc.). This benchmark would enable the acquirer to get an estimate of the value of the data with a more direct metric.

\textbf{Iterative Data Acquisition:} This work lays the design foundation of data acquisition benchmark by focusing on the one-shot acquisition strategies, i.e., first observing all available information in the data market and then determining what  to purchase once.
This captures several real-world applications, but many ML use cases are iterative in nature. Hence, the data acquisition process involves multiple iterations, too. For example, to train a health care assistant, one might first purchase a few thousand anonymous electronic health record data, and  then realize the shortage of data on Asian patients. After gaining these new data and retraining the model, she/he may notice the need for elderly or female data.  \textit{Iterative data acquisition} raises many interesting questions: how to allocate the budget among different iterations/rounds? How to leverage purchased datasets to help decide which new datasets to buy? 
And how to balance exploration and exploitation in an iterative acquisition process? 

\textbf{Data Labeling Selection:} Data labeling is important as many machine learning techniques are supervised. An alternative benchmark could focus on data labeling where data providers sell data labeling services instead of the raw datasets. This challenge is tailored for dataset acquirers to answer the question: given a fixed budget, how should an acquirer decide which data providers to query for the data labels, and how many labels to query from their unlabeled datasets?  


\textbf{Mechanism Design for Data Transactions:}
So far all challenges are tailoring to dataset acquirers. What does a data acquisition challenge look like from the perspective of data providers? Perhaps the most important question for providers is \textit{how to design an effective mechanism to sell their datasets. How to enable quantitative measures to enable acquirers the tool to evaluate how useful a dataset is}. At the same time, the evaluation mechanism must also ensure that the acquirer cannot infer individual data points in the provider’s dataset. How should the price of a dataset be determined to maximize revenue? 

\textbf{Dynamic Data Acquisition:}
This work presents the data acquisition benchmark design by assuming a fixed value function for data samples in static datasets on the marketplace. While the static dataset assumption represents certain real-world use cases, in many ways, machine learning datasets are dynamic in nature. For example, real-time data is constantly curated to capture evolving user interests or current events in modern deep learning recommender algorithms~\cite{Zhao:ISCA22}. Another example is federated learning, where data samples are continuously generated by a large pool of distributed client devices. Interesting opportunity arise for such dynamic, highly distributed machine learning environment --- \textit{what should a marketplace look like for data aggregation through federated learning? How should data value be specified to incentivize data sharing through federated learning participation?} 

\subsection{A Common Data Format}

The data acquisition benchmark design is our first step towards a consistent evaluation mechanism to assess and differentiate value of data. 
However, working with machine learning datasets on existing marketplaces is needlessly hard because each dataset comes with its unique file organization. 
The data format fragmentation across datasets on the marketplace and the lack of metadata tailoring to the datasets is a practical challenge faced by realistic data acquisition solutions. 

To enable effective data acquisition at-scale, we need standard data formats for machine learning. 
When data formats and metadata are standardized across datasets in a marketplace, evaluating the value add-on of new datasets is easier for data acquirers. It will also accelerate the development of data acquisition algorithms -- a key contribution of this work. Finally, it improves data quality and reduces the ever-increasing storage cost for AI data. 
We believe a common data format is key to propel the field and an enabling factor to effective data acquisition decisions.

\section{Related Work}

\textbf{Active Learning:} Active learning deals with the problem of iteratively selecting data points from a large (usually unlabeled) data pool (to be labeled)~\cite{settles2009active, zheng2002active}. It is based the setup that the ML model developer do have access to the full unlabeled data pool. This problem setup is not applicable to data acquisition from real life data marketplaces as the full data is not visible to the acquirer.

\textbf{Data Acquisition:} Existing research on data acquisition is not reflective of the real data markets.
For example one study proposes a data purchase algorithm for ML model training where the data is labeled and the price per data instance is fixed~\cite{li2021data}. The work relies on iterative data sampling and purchase however as we discussed earlier, in some datamarkets datasets are sold as bulk instead of individual samples.

Other work suggested \textit{Try Before You Buy} approach provides an efficient algorithm for evaluating a list of datasets for ML and then deciding which one to buy~\cite{andres2022try}. However it relies on full access to the datasets, which is not reflective of the real data markets.

An alternative way to solve the problem of limited information share between providers and acquirers was proposed through a platform that incentives the providers to share their data in exchange for rewards~\cite{fernandez2020data}. Whether such a platform can be effective or not in real markets is not clear.

\textbf{Data Pricing for ML:}
There is an increasingly growing interest in analyzing and designing data pricing mechanisms for ML~\cite{chen2019towards, liu2021dealer, chen2022selling, agarwal2019marketplace, pei2020survey, cong2022data}. 
For example, ~\cite{agarwal2019marketplace} designs a data marketplace for exchanging ML training data with a focus on fairness. ~\cite{chen2019towards} proposes a model-based pricing mechanism which offers arbitrage-freeness and revenue optimality.
Furthermore, ~\cite{liu2021dealer} integrates this mechanism with differential privacy. We refer interested readers to comprehensive surveys on this topic~\cite{pei2020survey, cong2022data}. Data pricing mechanism designs often aim at optimizing utility of data sellers, while our benchmark focuses on aiding the data acquirers in the existing marketplaces.

\textbf{Data Valuation:} Data valuation studies the contribution of individual data points to the trained ML models~\cite{jiang2023opendataval}. Among others, Shapley value~\cite{ghorbani2019data} has become the de facto approach to quantify data values.
Several techniques have been developed to make it more computationally efficient on specific learning models~\cite{jia2019towards,jia2019efficient,kwon2021efficient}, extend it to take statistical aspects of data into account~\cite{ghorbani2020distributional}, and twist it for noise reduction~\cite{kwon2021beta}. 
Existing data valuation techniques require white-box access to all data points, while data acquirers need to access data values before observing them.

%% file: 6_conclusion.tex
This paper presents a comprehensive study on the challenges and opportunities in data acquisition for ML systems. We highlight the lack of consistent methodologies and platforms offering detailed information about datasets, transparent pricing, and standardized data formats. To address these issues, we introduce the DAM benchmark, a model designed to optimize the interaction between data providers and acquirers.
Our analysis of the submitted strategies for the DataPerf benchmark underlines the need for effective data acquisition strategies in ML. Alternative benchmark designs and open problems are further discussed. 
We hope this paper lays a foundation for data acquisition in data-centric AI and stimulates researchers to tackle important challenges in the area.

%% file: _appendix.tex
\section{Detailed Performance of All Evaluated Strategies}

\begin{table}[htb!]
  \centering
  \caption{The performance of different allocation strategies on the five distinct markets.}
    \begin{tabular}{c||c|c|c|c|c}
    \hline
    Allocation Strategy &
      Market 1 &
      Market 2 &
      Market 3 &
      Market 4 &
      Market 5
      \bigstrut\\
    \specialrule{.12em}{.05em}{.05em}
    Strategy-All &
      74.41 &
      74.16 &
      73.86 &
      73.95 &
      73.44
      \bigstrut\\
    \hline
    Strategy-20\% &
      75.56 &
      74.38 &
      74.85 &
      75.04 &
      74.97
      \bigstrut\\
    \hline
    Strategy-40\% &
      75.46 &
      73.86 &
      75.06 &
      74.97 &
      75.36
      \bigstrut\\
    \hline
    Strategy-60\% &
      74.94 &
      71.82 &
      74.48 &
      74.92 &
      74.96
      \bigstrut\\
    \hline
    Strategy-80\% &
      74.1 &
      71.05 &
      73.96 &
      75.07 &
      73.01
      \bigstrut\\
    \hline
    Strategy-Single-1 &
      75.27 &
      71.55 &
      72.38 &
      74.64 &
      72.38
      \bigstrut\\
    \hline
    Strategy-Single-2 &
      73.21 &
      72.14 &
      72.14 &
      72.02 &
      72.14
      \bigstrut\\
    \hline
    Strategy-Single-3 &
      52.01 &
      71.01 &
      71.01 &
      32.37 &
      71.01
      \bigstrut\\
    \hline
    Strategy-Single-4 &
      71.65 &
      56.03 &
      75.99 &
      46.82 &
      75.99
      \bigstrut\\
    \hline
    Strategy-Single-5 &
      74.49 &
      74.95 &
      73.72 &
      77.43 &
      76.05
      \bigstrut\\
    \hline
    Strategy-Single-6 &
      75.45 &
      73.04 &
      74.57 &
      73.76 &
      74.58
      \bigstrut\\
    \hline
    Strategy-Single-7 &
      73.03 &
      70.46 &
      72.99 &
      74.89 &
      72.99
      \bigstrut\\
    \hline
    Strategy-Single-8 &
      50.5 &
      71.14 &
      70.1 &
      71.77 &
      43.6
      \bigstrut\\
    \hline
    Strategy-Single-9 &
      73.52 &
      73.32 &
      73.91 &
      75.88 &
      76.8
      \bigstrut\\
    \hline
    Strategy-Single-10 &
      74.28 &
      73.32 &
      78.36 &
      76.96 &
      76.51
      \bigstrut\\
    \hline
    Strategy-Single-11 &
      72.24 &
      71.58 &
      56.19 &
      73.96 &
      74.75
      \bigstrut\\
    \hline
    Strategy-Single-12 &
      71.81 &
      74.77 &
      75.86 &
      76.17 &
      76.99
      \bigstrut\\
    \hline
    Strategy-Single-13 &
      70.78 &
      72.84 &
      76.68 &
      74.91 &
      75.25
      \bigstrut\\
    \hline
    Strategy-Single-14 &
      77.63 &
      73.88 &
      71.73 &
      72.51 &
      48.08
      \bigstrut\\
    \hline
    Strategy-Single-15 &
      72.61 &
      76.36 &
      77.59 &
      77.25 &
      77.71
      \bigstrut\\
    \hline
    Strategy-Single-16 &
      75.93 &
      77.94 &
      76.14 &
      74.21 &
      75.05
      \bigstrut\\
    \hline
    Strategy-Single-17 &
      69.27 &
      76.59 &
      44.26 &
      73.21 &
      72.79
      \bigstrut\\
    \hline
    Strategy-Single-18 &
      74.45 &
      70.86 &
      68.92 &
      69.26 &
      71.95
      \bigstrut\\
    \hline
    Strategy-Single-19 &
      49.53 &
      74.12 &
      74.58 &
      74.48 &
      53.3
      \bigstrut\\
    \hline
    Strategy-Single-20 &
      76.48 &
      77.02 &
      74.48 &
      78.06 &
      76.21
      \bigstrut\\
    \hline
    Strategy-RFE &
      74.93 &
      76.76 &
      76.72 &
      72.54 &
      75.3
      \bigstrut\\
    \hline
    Strategy-CoFR &
      73.45 &
      76.96 &
      73.75 &
      76.58 &
      77.11
      \bigstrut\\
    \hline
    Strategy-$\ell_1/\ell_2$ &
      47.39 &
      74.95 &
      73.35 &
      77.16 &
      76.31
      \bigstrut\\
    \hline
    Strategy-$\ell_\infty$ &
      48.03 &
      75.1 &
      73.4 &
      77.21 &
      76.18
      \bigstrut\\
    \hline
    \end{tabular}%
  \label{tab:performance}%
\end{table}%

\begin{figure}[t]
  \centering
    \centering
    \vfill
    \includegraphics[width=1\textwidth]{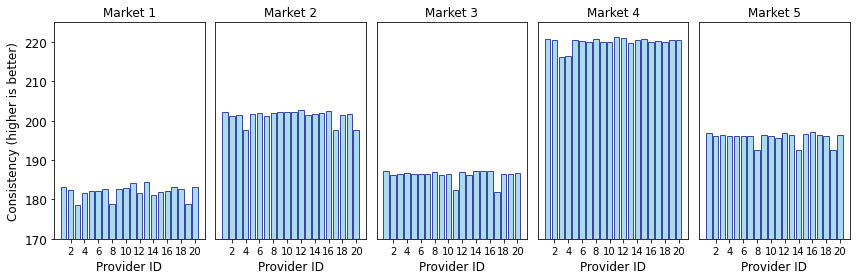}
    \caption{Consistency between acquier's data and data from each provider evaluated on all 768 features. Across all data markets, discrepancies between data providers on this measure is rather low. It indeed reveals the few problematic data providers with a much consistency score, but is hard to differentiate between other data providers or identify high quality providers.}
    \vfill
  \label{fig:sol2}
\end{figure}

\begin{figure}[t]
  \centering
    \centering
    \vfill
    \includegraphics[width=1\textwidth]{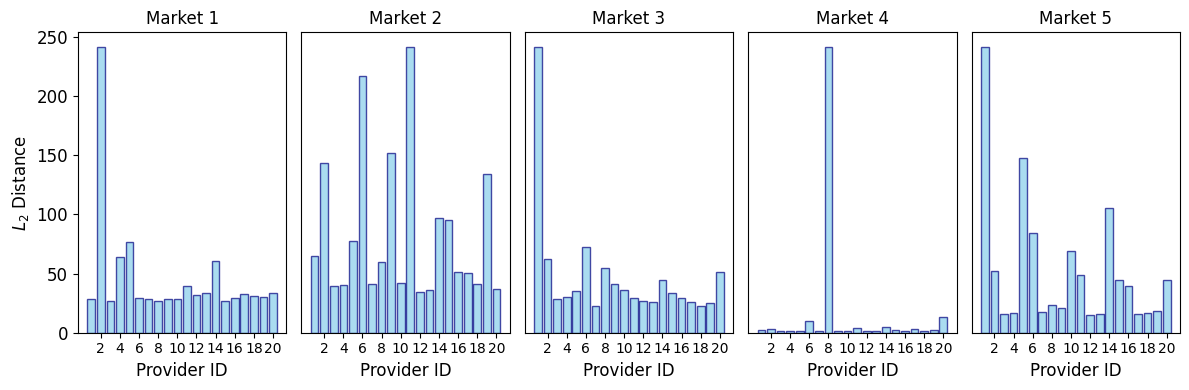}
    \includegraphics[width=1\textwidth]{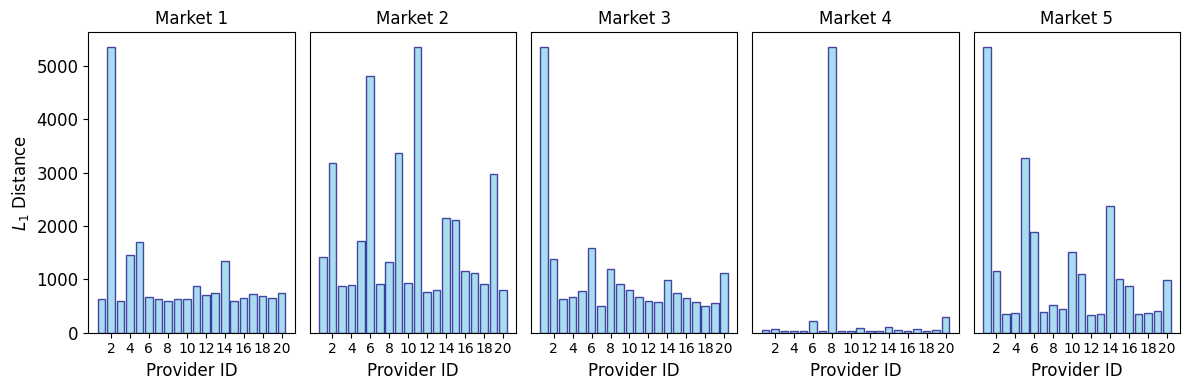}
    \includegraphics[width=1\textwidth]{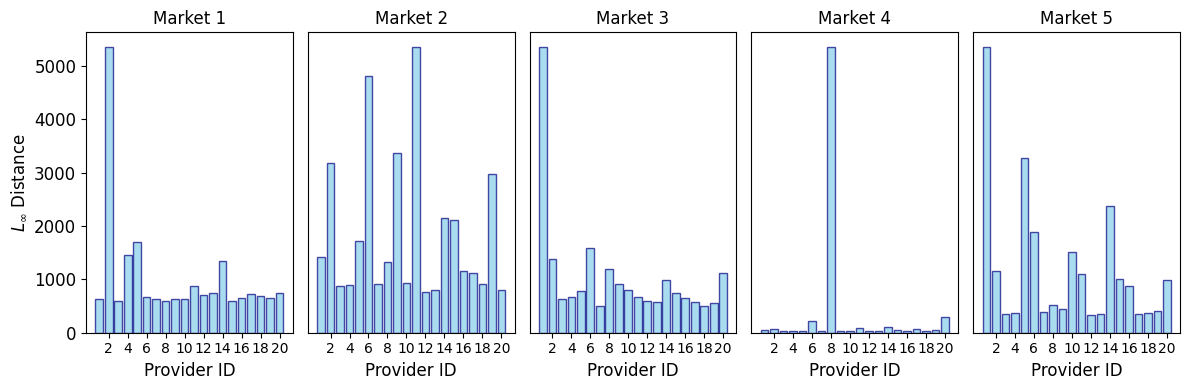}
    \caption{Top to bottom rows: $L_2$/$L_1$/$L_\infty$ distance between  coefficients of the model trained on acquier’s data and the feature relevance for data from each provider evaluated on all 768 features. The trend observed in each row are highly similar with the exact rankings slightly different. This measure provides strong discrimination between different data providers. Yet in market 1, the identified best data provider \#3 results in rather poor performance. This provider can be effectively identified as low quality in Strategy-RFE and Strategy-CoFR.}
    \vfill
  \label{fig:sol31}
\end{figure}